\pdfoutput=1

\documentclass[11pt]{article}

\usepackage[final]{acl}

\usepackage{times}
\usepackage{latexsym}

\usepackage[T1]{fontenc}

\usepackage[utf8]{inputenc}

\usepackage{microtype}

\usepackage{inconsolata}

\usepackage{graphicx}
\usepackage{enumitem}
\usepackage{multirow}
\usepackage{float}
\usepackage{amssymb}
\usepackage{amsmath}
\usepackage{tabularx}

\title{You Have Thirteen Hours in Which to Solve the Labyrinth:\\Enhancing AI Game Masters with Function Calling}

\author{Jaewoo Song\thanks{The first author conducted this work as a research assistant during his master's program at the University of Pennsylvania.} \\ Amazon Web Services \\ \tt{jaewoos@amazon.com}
        \And  
        Andrew Zhu \\ University of Pennsylvania \\ \tt{andrz@seas.upenn.edu}
        \And
        Chris Callison-Burch \\ University of Pennsylvania \\ \tt{ccb@seas.upenn.edu}}

\begin{document}
\maketitle
\begin{abstract}
Developing a consistent and reliable AI game master for text-based games is a challenging task due to the limitations of large language models (LLMs) and the complexity of the game master's role. This paper presents a novel approach to enhance AI game masters by leveraging function calling in the context of the table-top role-playing game "Jim Henson's Labyrinth: The Adventure Game." Our methodology involves integrating game-specific controls through functions, which we show improves the narrative quality and state update consistency of the AI game master. The experimental results, based on human evaluations and unit tests, demonstrate the effectiveness of our approach in enhancing gameplay experience and maintaining coherence with the game state. This work contributes to the advancement of game AI and interactive storytelling, offering insights into the design of more engaging and consistent AI-driven game masters.
\end{abstract}

\section{Introduction}
Imagine a world where the power of storytelling meets the ingenuity of artificial intelligence, giving rise to game masters that can weave captivating narratives and adapt to the players' choices in real-time. This is the vision that drives our research into enhancing AI game masters for table-top role-playing games (TTRPGs). However, the realization of this vision is hindered by the limitations of current large language models (LLMs) and the inherent complexity of the game master's role \cite{liapis2014computational}.

The popularity of LLMs sparked a wave of research on AI game masters \cite{hua2020playing,callison2022dungeons,zhou2022cast,zhu2023fireball,ang2023fable,triyason2023exploring}, but the challenge of maintaining consistency and coherence with the game state across multiple turns remains largely unaddressed. Indeed, using an LLM for a game master allows a variety of inputs and diverse narratives, unlike the traditional keyword-matching approach that requires rigid input commands and fixed outputs. However, an LLM-based game master is prone to going off the rails with respect to game rules and flow due to its unpredictability and limitation in performing game-specific functionalities. This is where our work comes in, proposing a novel approach that leverages function calling \cite{schick2024toolformer, li2024large} to provide fine-grained controls to the AI game master, enabling it to generate narratives that are not only engaging but also consistent with the game rules and state.

The main contributions of this paper are as follows:
\begin{itemize}
\setlength\itemsep{0.2em}
\item We present a methodology for enhancing AI game masters by integrating function calling, which allows for game-specific controls and state management.
\item We implement a simulation of the TTRPG ``Jim Henson's Labyrinth: The Adventure Game'' to evaluate the effectiveness of our approach in a realistic game setting.
\item We conduct human evaluations and unit tests to assess the impact of function calling on the narrative quality, state update consistency, and overall gameplay experience.
\item We provide insights and guidelines for designing more engaging and consistent AI-driven game masters, contributing to the advancement of game AI and interactive storytelling.
\end{itemize}

\begin{figure*}
    \includegraphics[width=\linewidth]{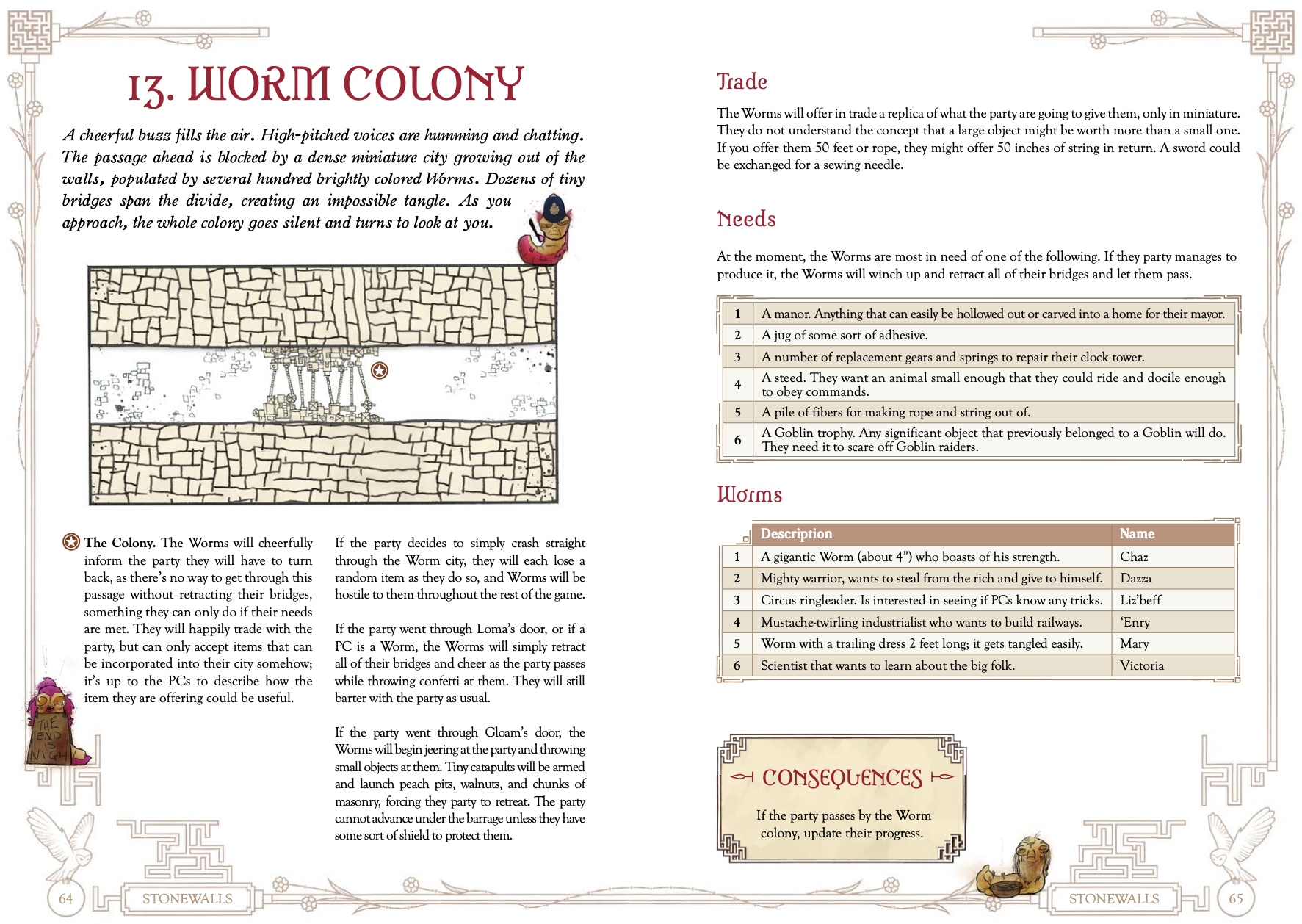}
    \caption{An example of a pre-written adventure location in ``Jim Henson's Labyrinth: The Adventure Game''.}
\end{figure*}

\section{Related Work}

The integration of AI in game design and development has been a topic of growing interest in recent years. In this section, we review the relevant literature on AI game masters, with a focus on the use of LLMs and function calling.
\subsection{AI Game Masters in TTRPGs}
One of the most actively researched TTRPGs in the domain of LLMs would be Dungeons \& Dragons (D\&D) (\citealp{gygax_arneson1974dnd}). \citet{callison2022dungeons} model D\&D as a dialogue challenge and experiment on the LLM's performance on the next utterance prediction and game state tracking, using the model to generate the next state. \citet{zhu2023fireball} present a dataset based on the ground-truth game states collected from real gameplay and test their effects on narration generation. 

\citet{zhu2023calypso} show that GPT-3 \cite{brown2020language} can be used as a game master's assistant to help brainstorm and create random encounters in D\&D. \citet{santiago2023rolling} use an LLM as a story-telling assistant in D\&D including a few generated story examples. Both these projects focus on using an LLM as an assistant, not a fully functional GM.
\

\subsection{LLMs with Function Calling}
The integration of function calling with LLMs has shown promising results in various domains. \citet{schick2024toolformer} propose a method for fine-tuning LLMs with API call annotations, enabling them to perform tasks such as question answering, calculation, and translation. \citet{li2024large} apply function calling to dialogue state tracking by mapping domains to functions and slot-value pairs to argument-value pairs.

In the context of games, \citet{volum2022craft} and \citet{wang2023voyager} leverage LLM-generated functions to perform actions in Minecraft. However, these projects focus on open-ended gaming agents rather than game masters, which arguably have more complex requirements and responsibilities.

Our work builds upon these previous studies by integrating function calling with LLMs to enhance AI game masters in the context of TTRPGs. We propose a methodology that allows for game-specific controls and state management, enabling more consistent and engaging gameplay experiences.

\section{Overview of Labyrinth}

To evaluate the effectiveness of our approach, we implement a simulation of the TTRPG ``Jim Henson's Labyrinth: The Adventure Game'' \cite{milton2020labyrinth} using the chat-based framework Kani \cite{zhu2023kani}. Labyrinth is a TTRPG inspired by the 1986 fantasy film ``Labyrinth,'' directed by Jim Henson.  In Labyrinth, players take on the roles of adventurers navigating a magical and treacherous maze filled with challenges and obstacles. The game master is responsible for describing the game world, controlling non-player characters (NPCs), and enforcing the game rules.
 Here are some key features of the game:
 
\begin{itemize}
\setlength\itemsep{0.2em}
    \item 
        System and Rules: The game is designed with newcomers in mind, and has a simpler rule set than D\&D. In the character creation process, players pick a class, with one character trait and one flaw (which affect their skill checks).
    \item 
        Dice-Based ``Tests'': The game primarily uses six-sided dice (d6) to determine the outcome whenever a character tries something that has a chance of failure. If the result is higher or equal to the difficulty number set by the GM, the character succeeds, otherwise they fail.  Relevant character traits cause dice to be rolled with advantage (rolling 2d6 and keeping the highest), and flaws cause dice to be rolled with disadvantage (2d6, keeping the lower).
    \item 
        Locations: The game includes pre-written adventures through a variety of locations in the Labyrinth.  Each location describes criteria that players must achieve in order to move on to the next location.  Most locations contain objects, NPCs and random tables which are used for initializing the scene or defining random encounters. 
    \item Time tracking: 
        The players are given 13 hours to reach the center of the Labyrinth and defeat the Goblin King. Any failure to pass the exit criteria or succeed in a certain task increments the clock.
\end{itemize}

\section{Labyrinth Game Simulation}

In our simulation of Labyrinth, the players create characters to explore the Labyrinth, and our AI agent takes on all of the responsibilities of the game master. More details in the implementation are in Appendix \ref{sec:appendixA}.

\subsection{Game State}
There are two types of game states in this system.

\begin{itemize}
\setlength\itemsep{0.2em}
    \item 
        The \textbf{scene state}, which represents the current state of the game world, including the scene description, NPCs, objects, and success/failure conditions.  The details of the scene state are elaborated in Appendix \ref{sec:appendixB}.
    \item
        The \textbf{player state} encompasses the specifications of each player character, such as their name, kin, persona, traits, flaws, and inventory. The details of the player state are written in Appendix \ref{sec:appendixC}.
\end{itemize}

These game states are initialized before each scene starts and can be updated during the game by the AI game master using the provided functions, discussed below. The scene state and player states are represented as Python objects or variables and are included in the input prompt for every generation by default.

\begin{figure}[t!]
  \includegraphics[width=\columnwidth]{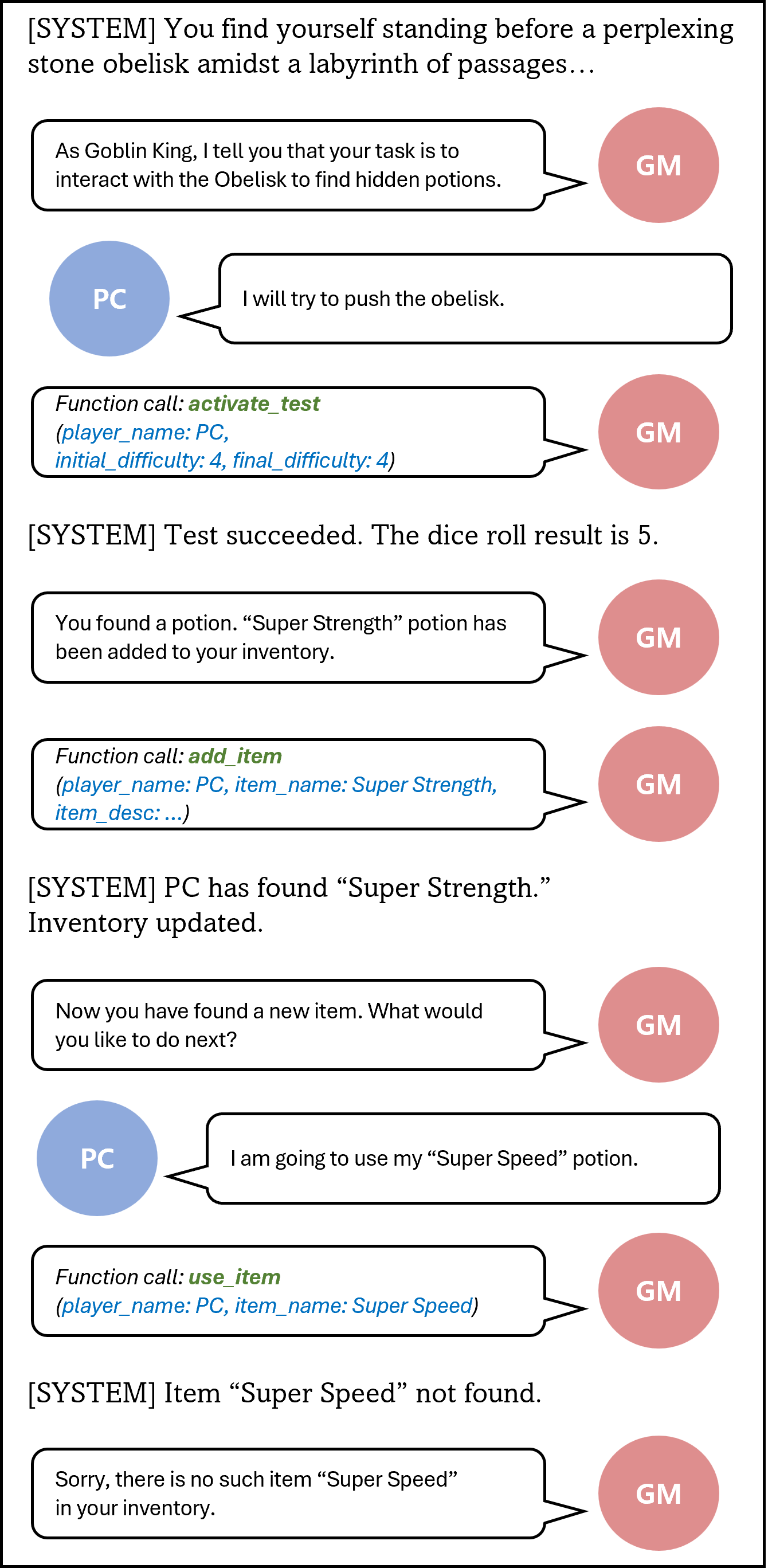}
  \caption{A dialogue between a player character (PC) and the game master (GM). This example shows how the AI GM uses different functions calls. Here, {\tt activate\_test} is called when the PC tries to do something challenging, {\tt add\_item} is called when the PC wants to add a new item to the inventory, and {\tt use\_item} is called when the PC tries to use an item in the inventory. The {\tt use\_item} fails because the PC asks to use a ``Super Speed potion" (which they do not have) instead of the ``Super Strength potion'' (which they do).  }
  \label{fig:sample-dialogue}
\end{figure}

\subsection{Rule Retrieval}
To maintain consistency with the game rules, we manually summarized the game rules by extracting the essential parts from the book. This rule summary, which consists of about 50 sentences, is injected into the prompt as a whole or a few sentences are retrieved according to the importance. More details are in Appendix \ref{sec:appendixA}.4.

\subsection{Dialogue history}
The dialogue history represents all of the turns in the chat, including player input, GM descriptions (output to the player), and function calls made by the GM.  A sample dialogue is given in Figure \ref{fig:sample-dialogue}. The chat messages in the history form the prompt to the game master depending on the pre-defined prompt design configurations, which are elaborated in Appendix \ref{sec:appendixA}.5.

\subsection{Function Types}
We define two types of functions in our game system:

\begin{enumerate}
\setlength\itemsep{0.2em}
    \item 
        \textbf{Dice roll function}: This function simulates the rolling of dice when a player attempts an action with a certain difficulty. For example, the {\tt activate\_test} function generates random numbers to mimic dice rolls and determines the success or failure of the action based on the outcome and the game rules. A dice roll function affects the game flow but does not directly modify the game state.
    \item 
        \textbf{State functions}: These functions directly modify the game state variables. For instance, the {\tt create\_npc} function adds a new NPC to the current scene, while the {\tt add\_item} function updates a player's inventory by adding a new item. State functions are essential for maintaining consistency between the game narrative and the underlying game state.
\end{enumerate}

 The list of all functions we used for this work and their specifications are attached in Appendix \ref{sec:appendixD}. And the examples of function definitions we use based on Kani's format are shown in Appendix \ref{sec:appendixE}.


\begin{figure*}[t]
  \includegraphics[width=1.0\linewidth]{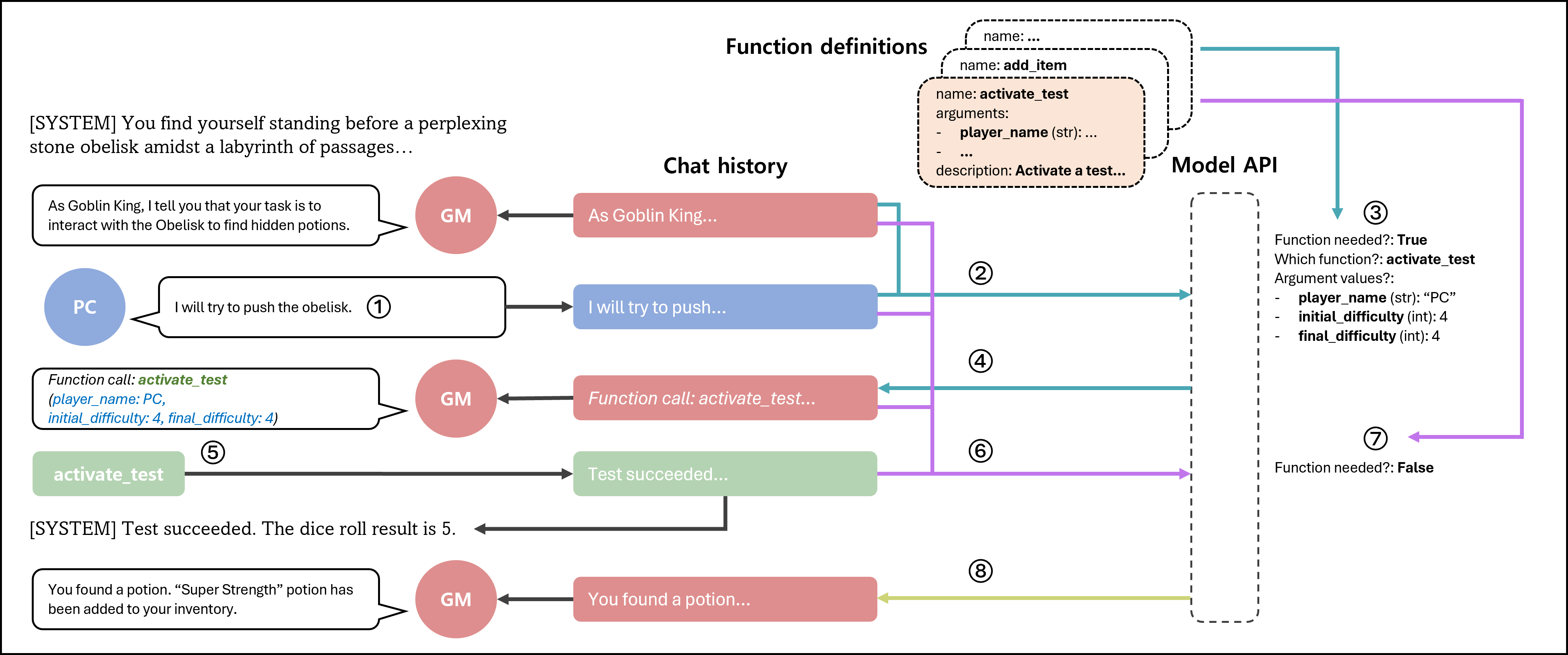} \hfill
  \caption {Steps involved in a single function call during gameplay: (1) The player message is added. (2) The game master agent makes a prompt with the chat history, game state, and function definitions. Then it requests an API call with the prompt. (3) The API determines whether to call a function or not. If it is needed, it chooses a function and parses/sets corresponding arguments from the chat history. (4) The API returns a function call message. It also comes with the parsed arguments. (5) When a function call is triggered, the pre-defined logic of it runs and returns the result message. (6) The game master calls the API with the updated chat history. (7) Again, it determines whether a function should be called or not. (8) If it is not needed, a normal response message is returned.}\label{fig:function-calling}
\end{figure*}

\subsection{Function Calling Process}
During gameplay, the AI game master determines when to call a function based on the current game context and the predefined function definitions. The function definitions, along with the chat history and game state, are passed to the language model as part of the input prompt. The model then selects the appropriate function to call and parses the necessary arguments from the dialogue context.

Figure \ref{fig:sample-dialogue} shows an example of multi-turn interaction between a player and the game master with functions. Normally, the game master generates a natural language response, but a function can be called anytime when it is necessary. Note that function calls can be sequential, where another function might be called after the previous one depending on the context or the result. Until the game master determines that no more functions or responses are needed (i.e. generates a stop token without a function call), the game master's turn continues.

Figure \ref{fig:function-calling} illustrates the steps involved in a single function call. The AI game master first generates a response based on the current game state and chat history. If a function call is required, the model selects the appropriate function and provides the necessary arguments. The function is then executed, updating the game state if necessary, and the result is appended to the chat history. This process continues until the game master determines that no further functions or responses are needed.

\section{Experimental Design}
In this section, we discuss the data collection and unit test procedures used to evaluate the performance of the AI game master.

\subsection{Data Collection}
To collect gameplay data, we simulate 24 game scenes from the Labyrinth game book using GPT-4\cite{achiam2023gpt} to play the roles of both the players and the game master. We create four player characters with diverse kins, traits, and flaws, and use the SentenceBERT \cite{reimers-2019-sentence-bert} model for retrieving relevant rule sentences based on the current input messages.
We test six different game master settings, varying the use of functions and game state management:

\begin{itemize}
\setlength\itemsep{0.2em}
    \item \textbf{FG-all}: Using all states and functions.
    \item \textbf{FG-dice}: Using all states and only dice roll function.
    \item \textbf{FG-states}: Using all states and only state functions.
    \item \textbf{FG-default}: Using all states, but no functions. All states stay the same after initialization.
    \item \textbf{FG-gen}: Using all states and the states are updated by GPT-4, not by functions.
    \item \textbf{DG}: Using no states and no functions. It sees the game states at the beginning, but they can be excluded due to the context size limit later.
\end{itemize}
Table \ref{tab:data-stats} summarizes the statistics of the collected gameplay data.

\begin{table}[t]
  \centering
  \begin{tabular}{p{0.35\textwidth}r}
    \hline
    \hline
    \multicolumn{2}{|c|}{\textbf{Labyrinth Gameplay Data}} \\
    \hline
    Total scripts    & 144  \\
    Total scenes    & 24    \\
    \hline
    Total utterances    & 4,937 \\
    Average utterances per script  & 34.28 \\
    \hline
    Total generated responses   & 1,021 \\
    Average generated responses & 7.09 \\
    \hline
    Total function calls    & 620 \\
    Average function calls per script & 8.64 \\
    \hline
    \hline
  \end{tabular}
  \caption{\label{tab:data-stats}
    The statistics of generated transcripts. The responses which have null content have been excluded. And the number of function calls has been calculated only from the scenes where the functions are used.
  }
\end{table}

\begin{table*}[h]
  \centering
  \begin{tabular}{p{2cm}|c|cccccc}
    \hline
    \hline
     \multirow{6}{*}{\textbf{Consistency}} &  \textbf{Setting}    & \textbf{FG-all}   & \textbf{FG-dice}  & \textbf{FG-states}    & \textbf{FG-default}   & \textbf{FG-gen}   & \textbf{DG} \\
    \cline{2-8}
    &   1   & \textbf{4.422}    & 3.867 & 3.711 & 3.356 & 3.756  & 3.600   \\
    &   2   & \textbf{4.333}    & 3.800 & 3.244 & 3.667 & 3.689 & 3.667   \\
    &   3   & \textbf{4.378}    & 4.000 & 3.311 & 3.467 & 3.578 & 3.689  \\
    \cline{2-8}
    &   Average & \textbf{4.378}    & 3.889 & 3.422 & 3.496 & 3.674 & 3.652 \\
    &   Total   & \textbf{4.388}    & 3.983 & 3.358 & 3.420 & 3.698 & 3.691 \\
    \hline
    \hline
  \end{tabular}
  \vspace{1em}

  \begin{tabular}{p{2cm}|c|cccccc}
    \hline
    \hline
     \multirow{6}{*}{\textbf{Reliability}} &  \textbf{Setting}    & \textbf{FG-all}   & \textbf{FG-dice}  & \textbf{FG-states}    & \textbf{FG-default}   & \textbf{FG-gen}   & \textbf{DG} \\
    \cline{2-8}
    &   1   & \textbf{3.922}    & 3.511 & 3.444 & 3.444 & 3.711 & 3.778 \\
    &   2   & \textbf{4.033}    & 3.556 & 3.267 & 3.644 & 3.711 & 3.711 \\
    &   3   & 3.878    & 3.600 & 3.222 & 3.311 & 3.600 & \textbf{3.889}\\
    \cline{2-8}
    &   Average & \textbf{3.944}    & 3.556 & 3.311 & 3.467 & 3.674 & 3.793 \\
    &   Total   & \textbf{3.949}    & 3.650 & 3.272 & 3.364 & 3.628 & 3.855 \\
    \hline
    \hline
  \end{tabular}
  \vspace{1em}

  \begin{tabular}{p{2cm}|c|cccccc}
    \hline
    \hline
     \multirow{6}{*}{\textbf{Interest}} &  \textbf{Setting}    & \textbf{FG-all}   & \textbf{FG-dice}  & \textbf{FG-states}    & \textbf{FG-default}   & \textbf{FG-gen}   & \textbf{DG} \\
    \cline{2-8}
    &   1   & \textbf{3.744}    & 3.267 & 3.600 & 2.667 & 3.022 & 3.333   \\
    &   2   & \textbf{3.722}    & 3.311 & 3.356 & 3.156 & 2.844 & 3.311 \\
    &   3   & \textbf{3.811}    & 3.422 & 3.444 & 2.933 & 2.889 & 3.422 \\
    \cline{2-8}
    &   Average & \textbf{3.759}    & 3.333 & 3.467 & 2.919 & 2.919 & 3.356 \\
    &   Total   & \textbf{3.765}    & 3.400 & 3.444 & 2.795 & 2.942 & 3.364 \\
    \hline
    \hline
  \end{tabular}

  \caption{\label{tab:human-eval}
    The human evaluation scores from 3 different samplings. A \textbf{bolded} score is the best score among each comparison. We also include the average of 3 samplings and the total average score of all evalauted responses without random sampling.
  }
\end{table*}

\subsection{Human Evaluation}
We recruit seven evaluators to assess the generated responses.  Each evaluator is assigned 12 game scripts (two scenes with six different settings) and asked to rate the responses on a Likert scale along three dimensions: \textit{consistency}, which is how well the model remains grounded in previous turns the game states, \textit{reliability}, which is how well the model follows the Labyrinth rules and role as GM, and \textit{interest}, which is how interesting the model's generation is. Full details of the human evaluation can be found in Appendix \ref{sec:appendixF}.

\begin{table}[th]
  \centering
  \begin{tabular}{p{0.03\textwidth}p{0.10\textwidth}p{0.10\textwidth}p{0.10\textwidth}}
    \hline
    \hline
        &   \textbf{FG-all} & \textbf{FG-states}    &   \textbf{FG-gen} \\
    \hline
    1   &   0.400    &   \underline{0.433}    &   0.233    \\
    2   &   0.417    &   \underline{0.483}    &   0.300    \\
    3   &   0.450    &   \underline{0.467}    &   0.267    \\
    \hline
    Avg    &   0.422    &   \textbf{0.461}    &   0.267    \\
    \hline
    \hline
  \end{tabular}
  \caption{\label{tab:unit-tests}
    Unit test results for state update correctness. \underline{Underlined} scores are the best score for each trial, and the \textbf{bold} score is the best among the average scores.
  }
\end{table}

\subsection{Unit Tests}
In addition to the gameplay data, we design 30 unit tests to compare the state update correctness between the different game master settings. Each test case consists of input states, input dialogue, and expected output states. The objective is to predict the output states correctly given the input states and dialogue.

The unit test cases are generated by augmenting the collected gameplay data, focusing on instances where state variables are changed after the game master generates a response. We use GPT-4 to paraphrase the dialogues while preserving the overall content and manually inspect the correctness of the state updates and the validity of the generated dialogues.

\section{Experimental Results}

\subsection{Human Evaluation Results}

Table \ref{tab:human-eval} presents the human evaluation scores for each setting, averaged over three different samplings. 

\subsubsection{Consistency}

For consistency, the \textbf{FG-all} setting, which uses both dice roll and state functions, outperforms the other settings. This demonstrates the effectiveness of integrating function calling in enhancing the consistency with the game progress. Appendix \ref{sec:appendixG} shows the statistical significance of \textbf{FG-all} calculated against other settings.

We found that the game easily fell into an undesired loop, where both players and game master infinitely wait for the dice roll without proceeding with anything. This "dice roll deadlock" hugely hurts the overall game experience and prevents the game master from following the game flow correctly. This is why \textbf{FG-dice} gets the second-best scores, showing the importance of the dice roll function to avoid this deadlock. Appendix \ref{sec:appendixH}.1 presents two different gameplay logs with and without the dice roll function.

\textbf{FG-default} and \textbf{FG-states} particularly fall behind in consistency. Especially, \textbf{FG-states} calls state functions too frequently, introducing new game states before resolving the previous challenges. This degrades the performance of the agent even worse. Appendix \ref{sec:appendixH}.3.1. shows one of the examples. Interestingly, \textbf{DG} shows the decent scores in consistency. We believe that \textbf{DG} is good at making up the dice result without the function since it does not leverage any states, allowing the game master to focus on the game rules better. We infer that the reason why \textbf{FG-gen} is also good at avoiding the deadlock is thanks to updating \verb|action_scene=True| variable. The action scene is activated when each player should take one dice roll action at a time to overcome an urgent circumstance. So it produces a signal like: "This is an action scene and you need to determine the result of each action!".

\subsubsection{Reliability}

For reliability, the \textbf{FG-all} also got the best scores in almost every sampling. This shows that using both function types helps the game master control and manage the game robustly. \textbf{DG} hits the nearly highest scores in reliability. By qualitative analysis, we found that \textbf{DG} tends to state the game rules explicitly and correct the player's trial more often. Again, this is an advantage of using only game rules without the extensive game states or function descriptions. This renders the game master seem more strict, whether its intervention is valid or not. Appendix \ref{sec:appendixH}.2 shows a few cases of how the game master corrected the user’s requests.

Unlike in consistency, \textbf{FG-dice} performs moderately in terms of reliability. While having the dice roll function mitigates the dice roll deadlock, that does not necessarily mean it is always beneficial. One of the feedbacks says that \textbf{FG-dice} often allows the player’s unrealistic moves too easily without determining whether they are valid or reasonable. Appendix \ref{sec:appendixH}3.2 includes a more detailed example. We conclude that the dice roll function and state functions intervene with each other, preventing excessive calls of certain functions and setting a proper balance during the game.

According to the appendix \ref{sec:appendixG}, \textbf{FG-all} shows a meaningful improvement compared to \textbf{FG-states} and \textbf{FG-default}. However, its performance is not statistically significant enough against \textbf{FG-gen} and \textbf{DG}. We believe the reliability metric has a lack of clarity and produces an unexpected bias even if the response is not good enough. This shows designing a more straightforward metric is essential for future works.

\subsubsection{Interest}

Overall, the settings with function calling (\textbf{FG-all}, \textbf{FG-dice}, \textbf{FG-states}) generate more specific and interesting responses. We see that functions are beneficial for improving the details and engagement of the output since the function message is integrated into the chat history and introduces additional context. Appendix \ref{sec:appendixG} presents that \textbf{FG-all} shows a worthwhile improvement in interest compared to \textbf{FG-default}, \textbf{FG-gen} and \textbf{DG}.

Interestingly, \textbf{DG} performs as great as \textbf{FG-dice} in interest among the settings that do not use function calling. While \textbf{DG} might introduce unrelated content during the game, this hallucination is actually considered interesting, regardless of its correctness.

\subsection{Unit Tests Results}
We conduct unit tests to evaluate the correctness of state updates for the \textbf{FG-all}, \textbf{FG-states}, and \textbf{FG-gen} settings. Table \ref{tab:unit-tests} presents the unit test results, showing the proportion of correctly predicted output states for each setting.

The \textbf{FG-states} setting consistently outperforms \textbf{FG-all} and \textbf{FG-gen} in the unit tests. This is because state functions can update the game state before the game master's turn is completed, whereas dice roll functions may cause the game master to consider the current challenge resolved without calling additional state functions.
However, it is important to note that the unit tests assume short-term interactions, and the superior performance of \textbf{FG-states} in this context does not necessarily translate to better performance in actual gameplay, where the lack of dice roll functions can lead to excessive function calls and disrupt the game flow. Appendix \ref{sec:appendixI} presents an example of a dialogue in one test case, and how the existence of a dice roll function causes the difference between the results of \textbf{FG-all} and \textbf{FG-states}.
\

\section{Conclusions}
In this research, we have demonstrated the effectiveness of integrating function calling with large language models (LLMs) to enhance the capabilities of AI game masters in the context of "Jim Henson's Labyrinth: The Adventure Game." Our experiments show that a combination of dice roll and state functions leads to the highest quality narratives and most engaging gameplay experiences, as evaluated by human raters. However, we also discovered that the optimal balance between these two types of functions is not always straightforward, with dice roll functions being crucial for smooth game flow and state functions being essential for maintaining consistency with the underlying game state.

Our work contributes to the growing body of research on the application of LLMs and function calling to game AI and interactive storytelling. By demonstrating the benefits and trade-offs of different function configurations in the context of a specific TTRPG, we provide valuable insights and guidelines for designing more engaging and consistent AI-driven game masters.

\section{Limitations and Future Work}

While our approach has shown promising results in the context of "Jim Henson's Labyrinth: The Adventure Game," there are several limitations to consider. First, the functions used in our study were specifically designed for this particular game, which may limit the generalizability of our findings to other TTRPGs or game genres. Future research could explore methods for automatically generating or adapting game-specific functions based on game manuals and rules, potentially enabling the application of our approach to a wider range of games.

Another limitation is the subjectivity inherent in human evaluations of the AI game master's performance. While we aimed to mitigate this by providing clear evaluation criteria and using multiple raters, the complexity and length of the game transcripts may have introduced some variability and bias in the ratings. Future work could investigate the use of more objective evaluation metrics and the potential for AI-assisted evaluation tools to handle longer and more complex interaction sequences.


Finally, while our work has implications for the broader field of game AI and interactive storytelling, these connections could be explored in more depth. Future research could investigate how the insights gained from our study of AI game masters in TTRPGs could be applied to other domains, such as video game NPCs, interactive fiction, or educational simulations. By continuing to bridge the gap between LLMs, function calling, and game AI, we can unlock new possibilities for creating engaging, adaptive, and immersive interactive experiences.


\bibliography{custom}

\clearpage

\appendix

\onecolumn
\section{Implementation details of Labyrinth}
\label{sec:appendixA}

\subsection{Overview}

\begin{figure}[H]
  \includegraphics[width=1.0\linewidth]{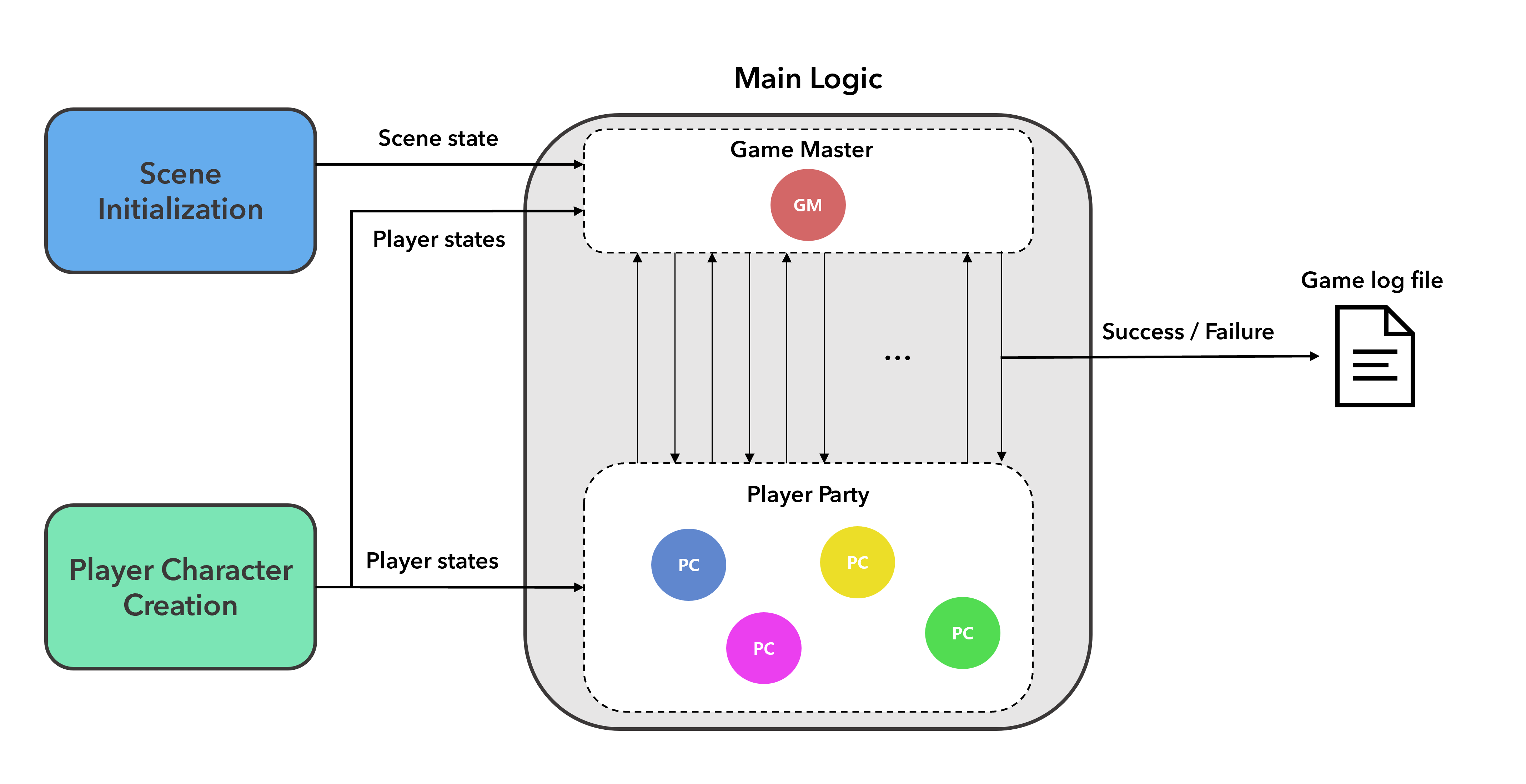} \hfill
  \caption {Overview of the game simulation. 1) The \textbf{scene initialization} step generates the starting scene state. 2) The \textbf{player character creation} step returns the created player states. 3) In the \textbf{main logic}, the actual game proceeds by the game master and player party. 4) The game terminates according to the \textbf{success/failure} condition and exports the \textbf{gameplay log} data.}
\end{figure}

\subsection{Scene initialization}

We manually parsed all scene text from the game book into a large JSON file. The data file has a list of JSON objects and each object represents the specifications of one scene, including the description, locations, notes, random tables, etc. However, these raw scene data are not formalized and have a low readability. Since most of the components are pure natural language texts, it is hard to parse a certain keyword or object from the scene. Also, the raw scene lacks details and the game master should improvise most of them in real-time during the game. To increase the consistency, we need more specific and informative initialization of scene components before the game, so that the model can keep track of the game states more easily.

In the scene initialization step, we convert a raw scene JSON object into a formalized scene state using GPT-4. Given a JSON input of the scene and the game rule summary, the model extracts, paraphrases, or creates the required scene state components. In more detail, the model generates the overall summary of the scene, the specifications of existing NPCs, the success condition of the scene, the failure condition of the scene, the intended game flow, and the environmental objects with their descriptions. All scene state components are organized in Appendix \ref{sec:appendixB}.

One of the most interesting components in Labyrinth is the random tables. The game master can use the randomly sampled entries from a table in various ways. For instance, the random samples can be new information or hints which might be useful for the players. Or they can be new challenges the players should overcome. The game master is also able to make random encounters such as new NPCs, objects or urgent circumstances. For scene initialization, we assume that a random table can be used for 1) nothing (only used during the game), 2) initializing NPCs, 3) initializing objects, or 4) initializing both NPCs and objects. We instructed GPT-4 to use the random tables for initializing a scene in the Chain-of-Thought approach, following these steps: First, GPT-4 determines which usage each random table falls into. If a table should be used for 2, 3, or 4, we ask GPT-4 how many samples should be retrieved from this table. If the raw scene input specifies the exact number, the model parses it. Otherwise, we just let the model decide a reasonable number. After that, we randomly sample the entries from the tables and feed the samples as the required conditions when GPT-4 generates other scene components, such as NPCs and objects. The tables used for initialization are removed and only those that have not been used remain in the scene state when the game starts.

\subsection{Player character generation}

Unlike the scene initialization, the player character is made by each human player. We parsed the "Creating character" section from the book and organized the available options a player can choose. Each player chooses one kin and each kin has the persona, default traits, flaws, or items accordingly. Then the player can set the name and goal freely. Finally, the player chooses one trait and flaw from the given list to complete the character creation. The created player information is converted into a player state in JSON format. All player state components are elaborated in Appendix \ref{sec:appendixC}.

\subsection{Rule injection}

There are two ways of injecting the rules into the prompt. First, \textbf{full injection} simply attaches the full rule summary. On the other hand, \textbf{retrieval injection} parses the top 5 relevant rule sentences given the current input messages and adds them to the prompt. In more detail, assume that there are $R$ rule sentences and $Q$ input messages. All of them are encoded into the vectors in size of $E$. With the rule matrix $G \in \mathbb{R}^{E \times R}$ and query matrix $M \in \mathbb{R}^{E \times Q}$, the cosine similarity matrix $C$ is calculated as follows:

\begin{equation}
  \label{eq:retrieval}
   C \in \mathbb{R}^{R \times Q}, \; C_{ij} = \frac{g \cdot m}{||g|| ||m||}, \; g = \begin{bmatrix} G_{1i} \\ G_{2i} \\ \vdots \\ G_{Ei} \end{bmatrix}, \; m = \begin{bmatrix} M_{1j} \\ M_{2j} \\ \vdots \\ M_{Ej} \end{bmatrix}
\end{equation}

Then, we take the max-pooled vector $C' \in \mathbb{R}^{R \times 1}$ to get the maximum score of each rule sentence. Finally, we pick the top 5 sentences with the highest scores and put them into the prompt.

\subsection{Prompt designs}

In this work, the user is able to set various combinations of prompt design approaches. By default, Kani provides a prompt construction algorithm to set the given messages to fit into the limited context window size. It excludes the least recent messages first until the total number of tokens in the messages is less than or equal to the context size, including the system instruction, function descriptions, and any other messages that are set to be always included. Besides that, we implemented the following variants in prompt design methodologies:
\begin{itemize}
    \item \textbf{Concatenation} \\
        \textbf{Simple concatenation} is just concatenating the messages in order, which is mostly used in a wide range of interactive AI systems. \textbf{Retrieval concatenation}, on the other hand, fetches the most relevant chat messages from the history given the current queries to process. This is actually the same mechanism as the retrieval rule injection, but the only difference is that the system attends to the utterances in the past chat history, not the rule sentences.
    \item \textbf{Maximum number of messages} \\
        The user can specify the \textbf{maximum number of messages} in the prompt. If it is not set, the system takes as many messages as possible within the context size. If a certain number is set, the system only takes a limited number of messages as specified.
    \item \textbf{Summarization} \\
        \textbf{Summarization} can be used in various ways. By default, summarization requires the \textbf{summarization period}, which indicates how frequently the chat history should be summarized. For example, if the period is 2, when every 2 interactions between the player party and the game master are finished, the game master summarizes the history so far and adds the result to the chat history. This summary can be used when the system concatenates the messages either with simple concatenation or retrieval concatenation. If the summarization period doesn't exist, the system summarizes the whole chat history every time it creates an input prompt. In this case, none of the other variants for prompt design matter.
    \item \textbf{Raw chat message handling} \\
        When the system leverages summarization, it can also either \textbf{remove the original chat messages} or \textbf{keep them}. In other words, the summary replaces the original chat messages that are used for summarization. In this way, the number of chat messages remaining in the history can be efficiently maintained.
\end{itemize}

\onecolumn
\section{Scene state details}
\label{sec:appendixB}

\begin{table}[H]
  \centering
  \begin{tabular}{p{0.20\textwidth}|p{0.25\textwidth}|p{0.55\textwidth}}
    \hline
    \textbf{Component}  & \textbf{Type} & \textbf{Description}  \\
    \hline
    \verb|chapter|  & \verb|str|    & The chapter name. \\
    \hline
    \verb|scene|  & \verb|str|  & The scene name. \\
    \hline
    \verb|scene_summary|  & \verb|list[str]|  & The brief summary of the current scene. Each string is one sentence. \\
    \hline
    \verb|npcs|  & \verb|dict[str, dict]|    & The initialized NPCs. Each key is an NPC name and the value is the specification, which is another dictionary. The specification includes kin, persona, goal, trait and flaw. \\
    \hline
    \verb|success_condition|  & \verb|str|    & The condition for the players to win this scene. \\
    \hline
    \verb|failure_condition|  & \verb|str|    & The condition for the players to fail this scene. \\
    \hline
    \verb|game_flow|  & \verb|list[str]|    & The intended game flow of the current game scene. Each string is one sentence. \\
    \hline
    \verb|environment|  & \verb|dict[str, str]|    & The environmental objects. Each key is an object name and the value is the description. \\
    \hline
    \verb|random_tables|  & \verb|dict[str, list[str]]|    &  The random tables. Each key is a table name and the value includes the string entries. \\
    \hline
    \verb|consequences|  & \verb|str|    & The consequences after finishing the scene. \\
    \hline
    \verb|is_action_scene|  & \verb|bool|    & Indication of whether the action scene is currently activated or not. \\
    \hline
  \end{tabular}
  \caption{\label{scene-state}
    The list of all components in the scene state. Note that each component is represented as the data type above in Python and put into the prompt after being converted into a flat string format.
  }
\end{table}

\section{Player state details}
\label{sec:appendixC}

\begin{table}[H]
  \centering
  \begin{tabular}{p{0.20\textwidth}|p{0.25\textwidth}|p{0.55\textwidth}}
    \hline
    \textbf{Component}  & \textbf{Type} & \textbf{Description}  \\
    \hline
    \verb|name|  & \verb|str|    & The name of the player. \\
    \hline
    \verb|kin|  & \verb|str|  & The kin of the player. \\
    \hline
    \verb|goal|  & \verb|str|  & The goal of the player. \\
    \hline
    \verb|traits|  & \verb|dict[str, str]|    & The traits of the player. Each key is a trait name and the value is the description. \\
    \hline
    \verb|flaws|  & \verb|dict[str, str]|    & The flaws of the player. Each key is a flaw name and the value is the description. \\
    \hline
    \verb|inventory|  & \verb|dict[str, str]|    & The inventory of the player. Each key is an item name and the value is the description. \\
    \hline
    \verb|additional_notes|  & \verb|list[str]|    & This adds an additional notes regarding the player character. This is something like: A player does something, add a new trait/flaw, etc. \\
    \hline
  \end{tabular}
  \caption{\label{player-state}
    The list of all components in the player state. Note that each component is represented as the data type above in Python and put into the prompt after being converted into a flat string format.
  }
\end{table}

\onecolumn
\section{List of all functions}
\label{sec:appendixD}

\begin{table}[H]
  \centering
  \begin{tabular}{p{0.27\textwidth}|p{0.28\textwidth}|p{0.25\textwidth}|p{0.1\textwidth}}
    \hline
    \textbf{Function}   & \textbf{Description}  & \textbf{Sub-tasks}    & \textbf{Category} \\
    \hline
    \verb|activate_test|    & It performs a dice roll when a player tries a test. The difficulty is set by the game master.    & If the player's attribute improves/hinders the test, two dice should be rolled and a larger/smaller value is picked. & Dice roll   \\
    \hline
    \verb|activate_action_scene|    & It starts an action scene.    & - & \multirow[t]{13}{*}{State}   \\
    \cline{1-3}
    \verb|terminate_action_scene|    & It ends an action scene.    & - &    \\
    \cline{1-3}
    \verb|create_npc|    & It creating a new NPC in the scene and add it into the scene state.   & A sub-LLM generates the NPC specifications in a JSON form. &    \\
    \cline{1-3}
    \verb|add_trait|    & It adds a new trait into a player's trait list.    & - &    \\
    \cline{1-3}
    \verb|add_flaw|    & It adds a new flaw into a player's flaw list.    & - &    \\
    \cline{1-3}
    \verb|add_item|    & It adds a new item into a player's inventory.    & - &    \\
    \cline{1-3}
    \verb|remove_trait|    & It removes a trait from a player's trait list.    & - &    \\
    \cline{1-3}
    \verb|remove_flaw|    & It removes a flaw from a player's flaw list.    & - &    \\
    \cline{1-3}
    \verb|remove_item|    & It removes an item from a player's inventory and leaving it in the environment.    & - &    \\
    \cline{1-3}
    \verb|use_item|    & It lets a player use an item in the inventory.    & If the item should be removed after usage, it is removed from the inventory. &    \\
    \cline{1-3}
    \verb|add_object|    & It adds a new object in the environment.    & - &    \\
    \cline{1-3}
    \verb|use_environment|    & It lets a player get access to an object in the environment.    & If the object is obtainable, the player can choose whether to take it or not. &    \\
    \cline{1-3}
    \verb|use_random_table|    & It samples some random entries from a random table. The results can introduce a new context or triggers another functions, such as \verb|create_npc| or \verb|add_object|.    & If the sampled entries or the table itself should be removed after usage, they are removed. &    \\
    \hline
  \end{tabular}
  \caption{\label{functions}
    The list of all functions which are used for this work. Each function has its name, description, category (dice roll / state) and sub-tasks depending on the design.
  }
\end{table}

\onecolumn
\section{Function definition examples}
\label{sec:appendixE}

\subsection{{\tt activate\_test} (Dice roll)}
\begin{figure}[H]
  \includegraphics[width=1.0\linewidth]{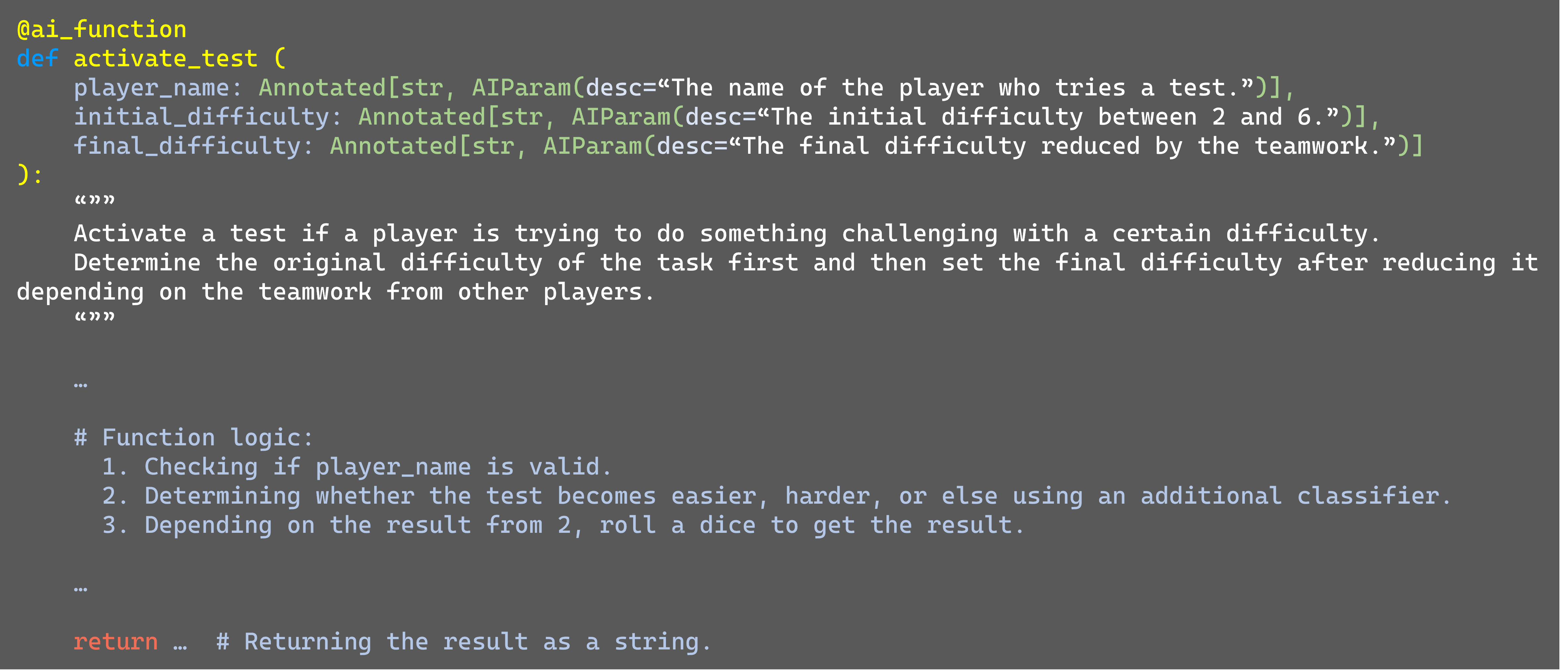} \hfill
  \caption {The definition of {\tt activate\_test} function which is called when a player should roll the dice. Each function definition has {\tt @ai\_function} annotation, function name, argument annotations, docstring, function logic, and the return value. In the actual implementation, the function logic part is a code block to perform the logic.}
\end{figure}

\subsection{{\tt create\_npc} (State)}
\begin{figure}[H]
  \includegraphics[width=1.0\linewidth]{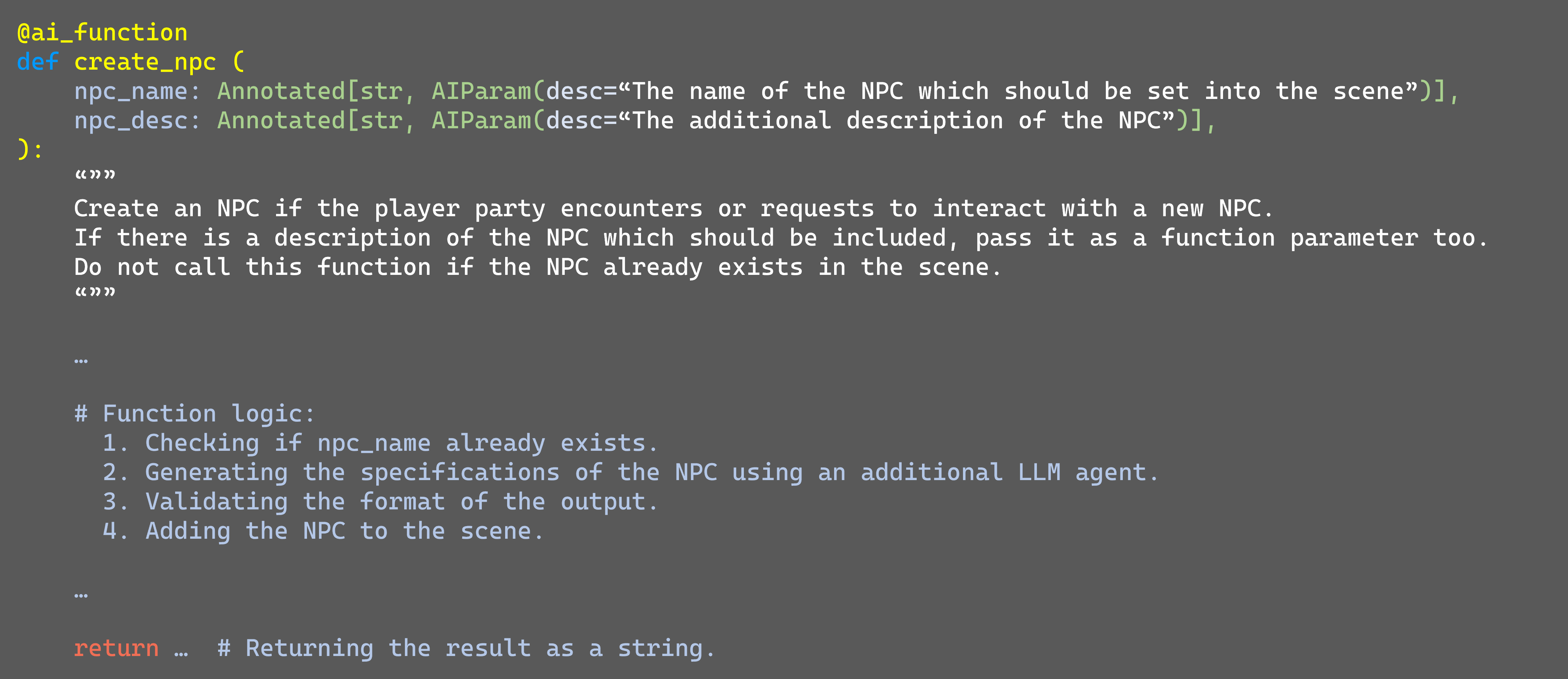} \hfill
  \caption {The definition of {\tt create\_npc} function which is called when a new NPC should be introdued during the scene. Likewise, in the actual implementation, the function logic part is a code block to perform the logic.}
\end{figure}

\onecolumn
\section{Survey questions}
\label{sec:appendixF}

\subsection{Consistency}
\textbf{How consistent is the target response to the current game progress, including the chat history and the game states?}

\begin{enumerate}
    \item The target response is consistent with the chat history between the players and the master so far.
        \begin{itemize}
            \item The model remembers the past interactions.
            \item The response is relevant to the player party's queries or requests.
        \end{itemize}
    \item The target response is consistent with the updates in the scene and players so far.
        \begin{itemize}
            \item The model acknowledges the existing components in the current scene, such as NPCs, objects, and random table entries.
            \item The model acknowledges the existing properties of the players, such as traits, flaws, and inventories.
        \end{itemize}
\end{enumerate}

\textreferencemark If the model output assumes or fakes up any non-existing components, ignore it for this question.This will be penalized in the reliability check question.
\\

(1=The model does not follow the progress at all, 3=The model makes a narration that is plausible but misses some components in the scene or players, 5=The model's response correctly follows the chat history while acknowledging the existing components in the states well too)

\subsection{Reliability}
\textbf{How well does the model control and manage the game reliably?}

\begin{enumerate}
    \item The game master fully understands the game and performs its task as a master correctly.
        \begin{itemize}
            \item The model keeps the general game rules in Labyrinth.
            \item The model understands the scene-specific rules, instructions, and specifications of the current scene and guides the players to proceed with the game as intended.
        \end{itemize}
    \item When a player tries to do something invalid, the game master rejects it robustly.
        \begin{itemize}
            \item The model rejects it when the player attempts to do something which cannot be performed by a player character or which is not the player's task.
            \item The model rejects it when the player tries to use a trait, flaw, or item which does not exist in the player.
            \item The model rejects it when the player tries to leverage or get access to non-existing objects, NPCs, or random tables.
        \end{itemize}
    \item Any unexpected behavior which might hurt the players' gameplay experience or make the game flow far from intended should be penalized.
\end{enumerate}

\textreferencemark Note that this metric does not evaluate the quality of the response. Even if the response looks perfect, it can contain an invalid content or the model might just let the player do an unallowed trial.
\\

(1=The model blatantly ignores the rules or is completely generous with the players' invalid moves, which makes the game go into a bad state, 3=The model gets some rules incorrect or accepts the players' some violations, but the game generally progresses as it should, 5=The model keeps the rules correctly and corrects the players' invalid or unacceptable behaviors)

\subsection{Interest}
\textbf{How interesting is the generated response?}

\begin{enumerate}
    \item The response describes the scene funny, entertaining and specific.
    \item The response makes the user engaged and immersed in the game.
\end{enumerate}

(1=The response is too bland, simple, or half-hearted, 3=The response is not highly entertaining, but at least it is not boring, 5=The response is so engaging and immersive that I wouldn't want to stop the game if I were a player)

\onecolumn
\section{Statistical significance}
\label{sec:appendixG}

\begin{table}[H]
  \centering
  \begin{tabular}{l|lllll}
    \hline
    &   \textbf{FG-dice}  & \textbf{FG-states} & \textbf{FG-default}    & \textbf{FG-gen}   & \textbf{DG} \\
    \hline
    \textbf{Consistency}    & 0.0835    & \textbf{0.0000}    & \textbf{0.0001}    & \textbf{0.0025}    & \textbf{0.0064} \\
    \textbf{Reliability}    & 0.2722    & \textbf{0.0094}    & \textbf{0.0338}    & 0.1806    & 0.7400 \\
    \textbf{Interest}    & 0.0965    & 0.1248    & \textbf{0.0000}    & \textbf{0.0005}    & \textbf{0.0427} \\
    \hline
  \end{tabular}
  \caption{\label{player-state}
    The p-values of \textbf{FG-all} against other settings in each metric after conducting t-tests. The \textbf{bolded} values are $p < 0.05$, which are considered statistically significant.
  }
\end{table}

\onecolumn
\section{Examples of the model's responses}
\label{sec:appendixH}

\subsection{Dice roll behaviors}

\subsubsection{With a dice roll function}

\begin{figure}[H]
  \includegraphics[width=1.0\linewidth]{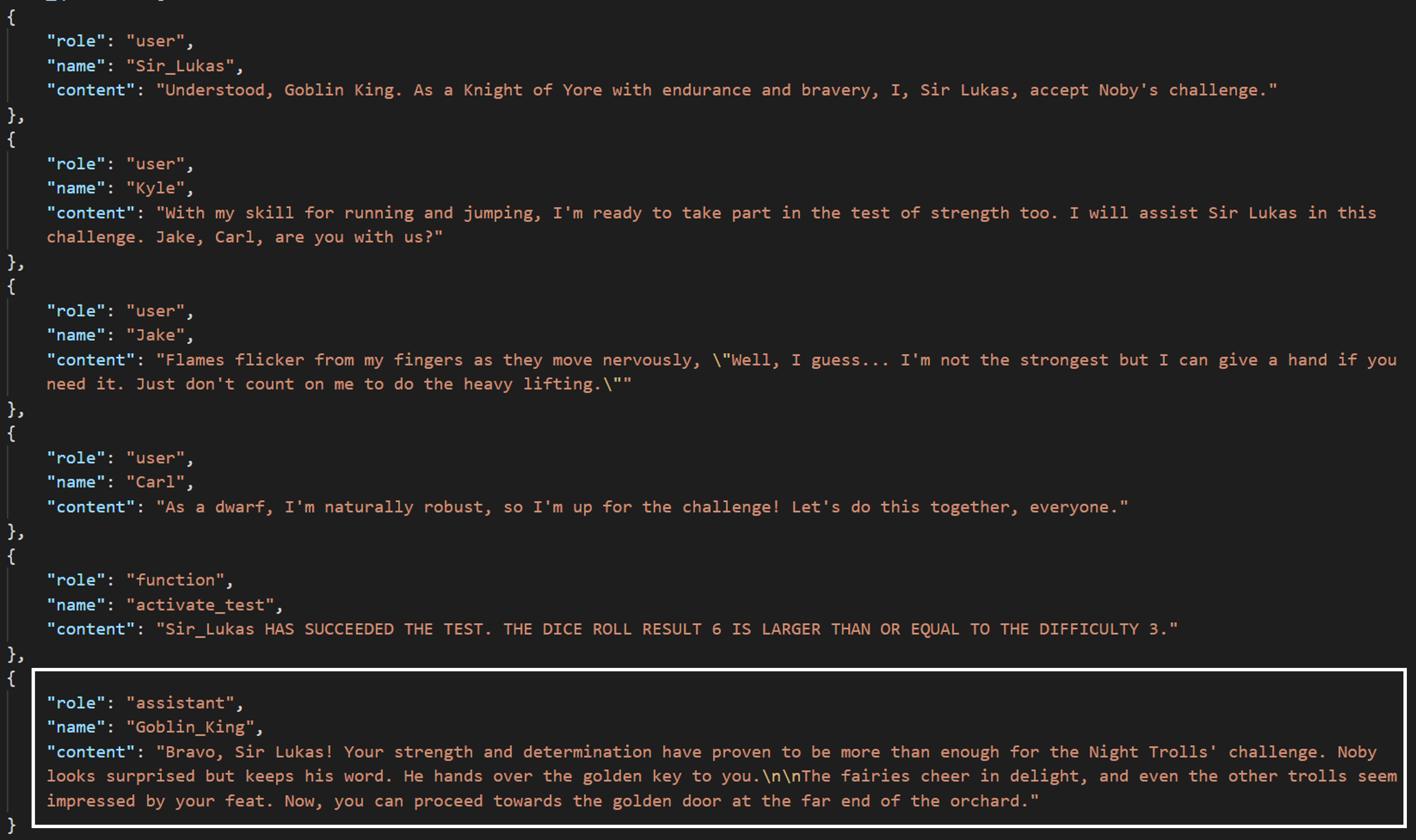} \hfill
  \caption {With the dice roll function, the game master properly notifies the result and let the game smoothly proceed.}
\end{figure}

\onecolumn

\subsubsection{Without a dice roll function}

\begin{figure}[H]
  \includegraphics[width=1.0\linewidth]{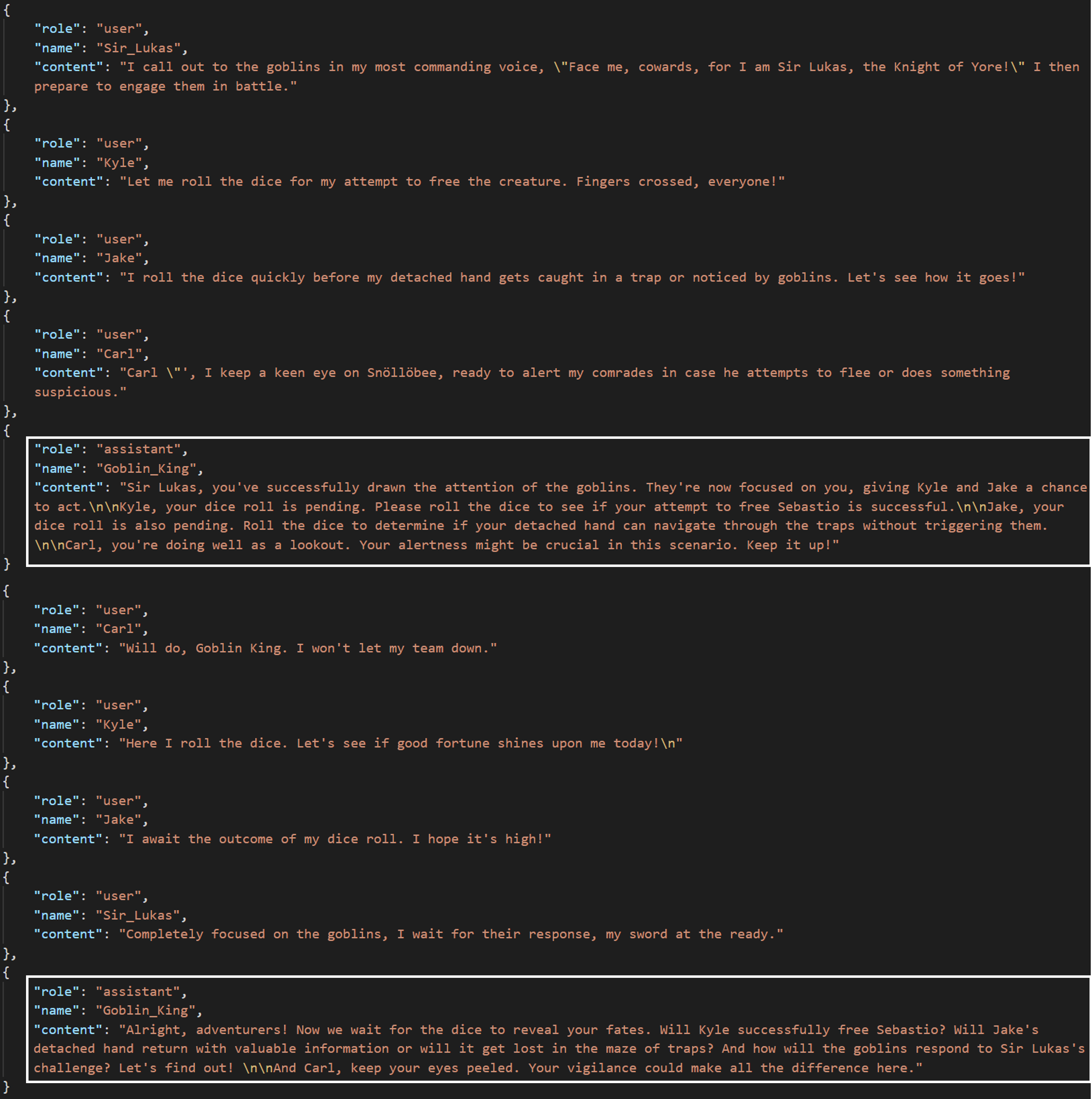} \hfill
  \caption {Without the dice roll function, the game master does not produce the dice roll result and keeps waiting for the result.}
\end{figure}

\onecolumn

\subsection{Controlling the behavior from the player}

\begin{figure}[H]
  \includegraphics[width=1.0\linewidth]{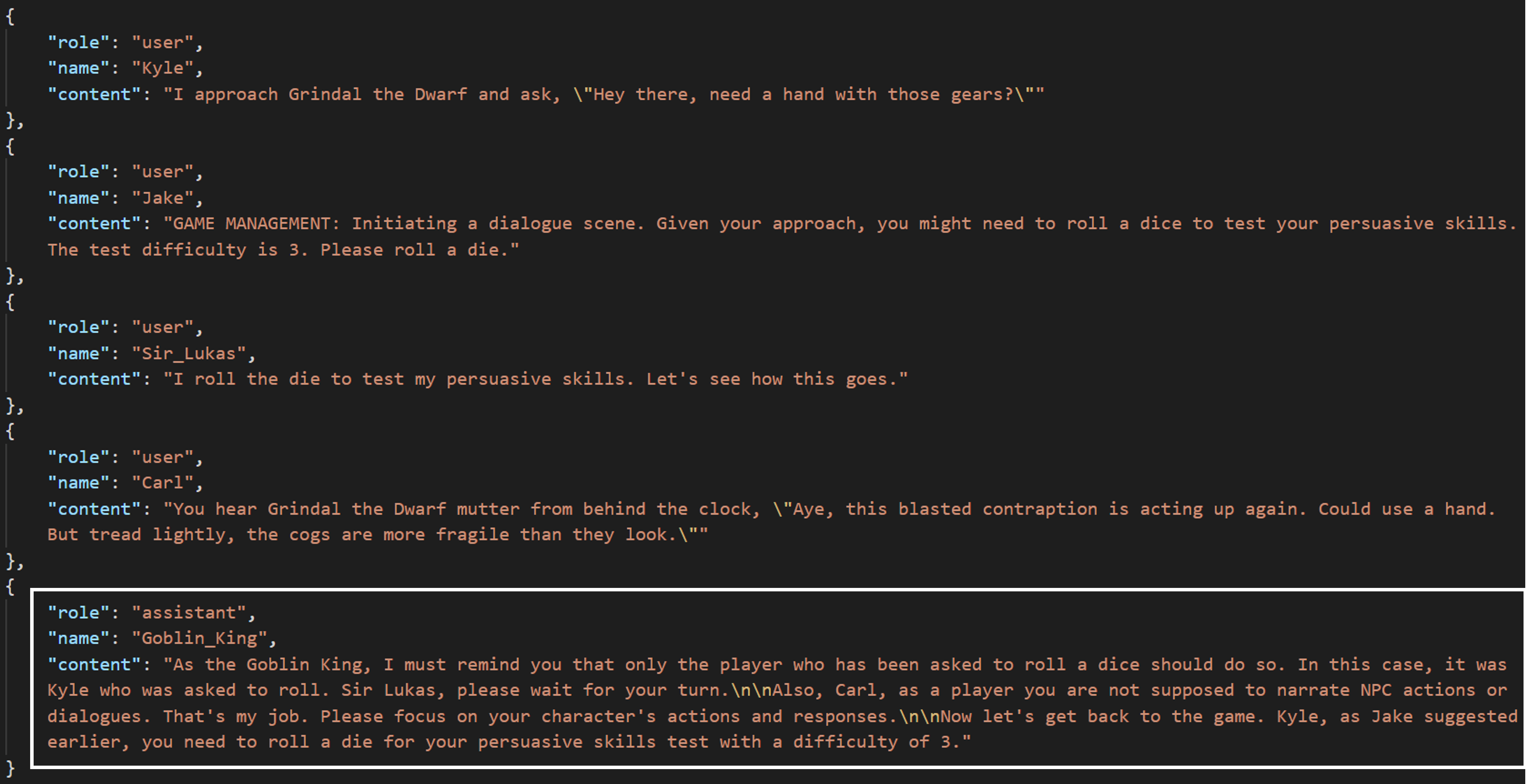} \hfill
  \caption {The game master properly corrects the player's behavior. In this case, it restrict a dice roll attempt from Sir Lukas and points out the player Carl's invalid move, which plays the game master's role. However, this is not perfect, since the master does not say anything about Jake's improper behavior, which is setting the difficulty.}
\end{figure}

\begin{figure}[H]
  \includegraphics[width=1.0\linewidth]{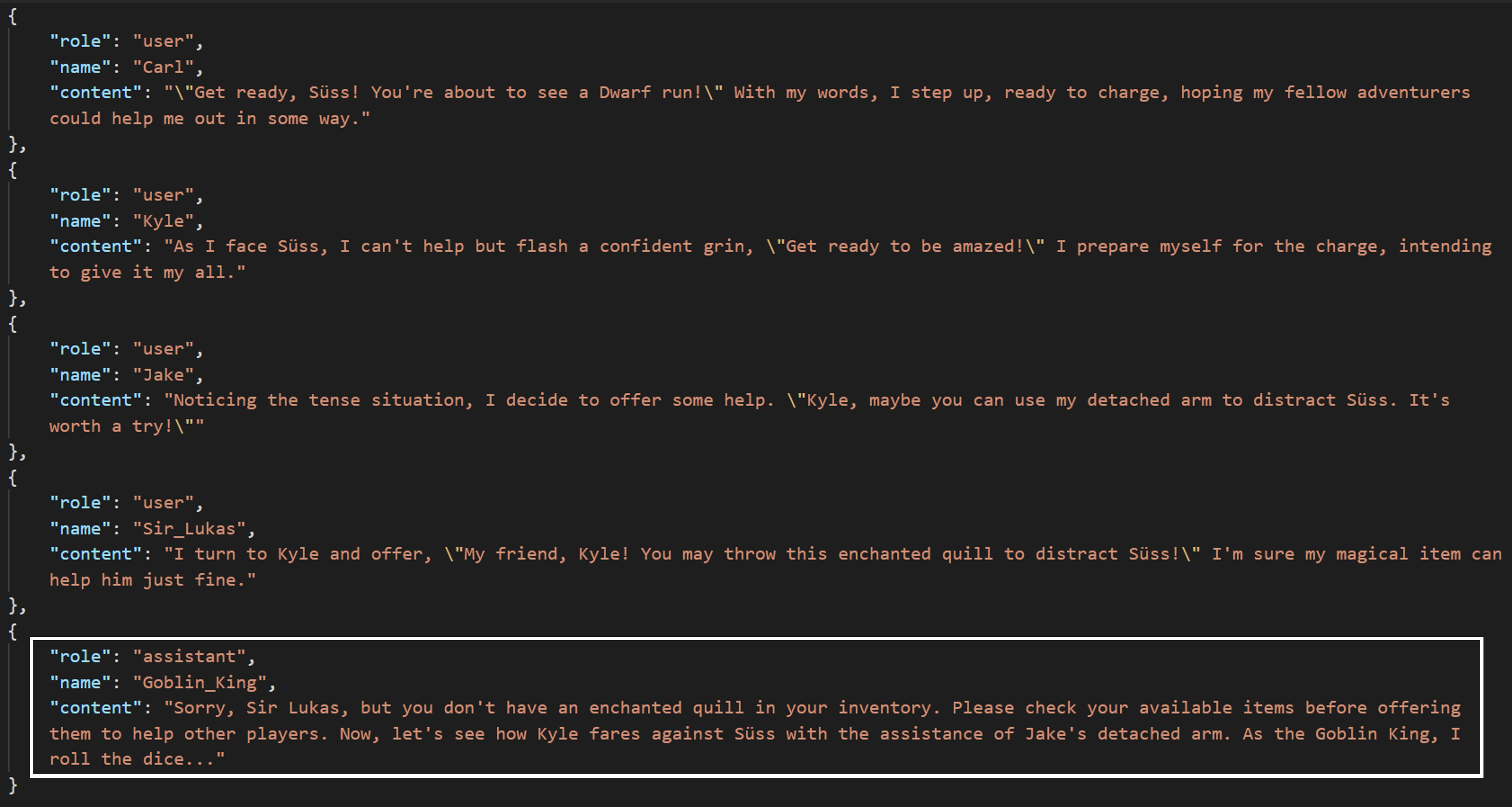} \hfill
  \caption {The game master corrects Sir Lukas's attempt to use Enchanted Quill. However, it is actually valid, since the player has the item. Still, the evaluator gave 5 points in reliability regardless of correctness.}
\end{figure}

\onecolumn
\subsection{Limitation of using one function type}

\subsubsection{Excessive calls of functions only with state functions}

\begin{figure}[H]
  \includegraphics[width=1.0\linewidth]{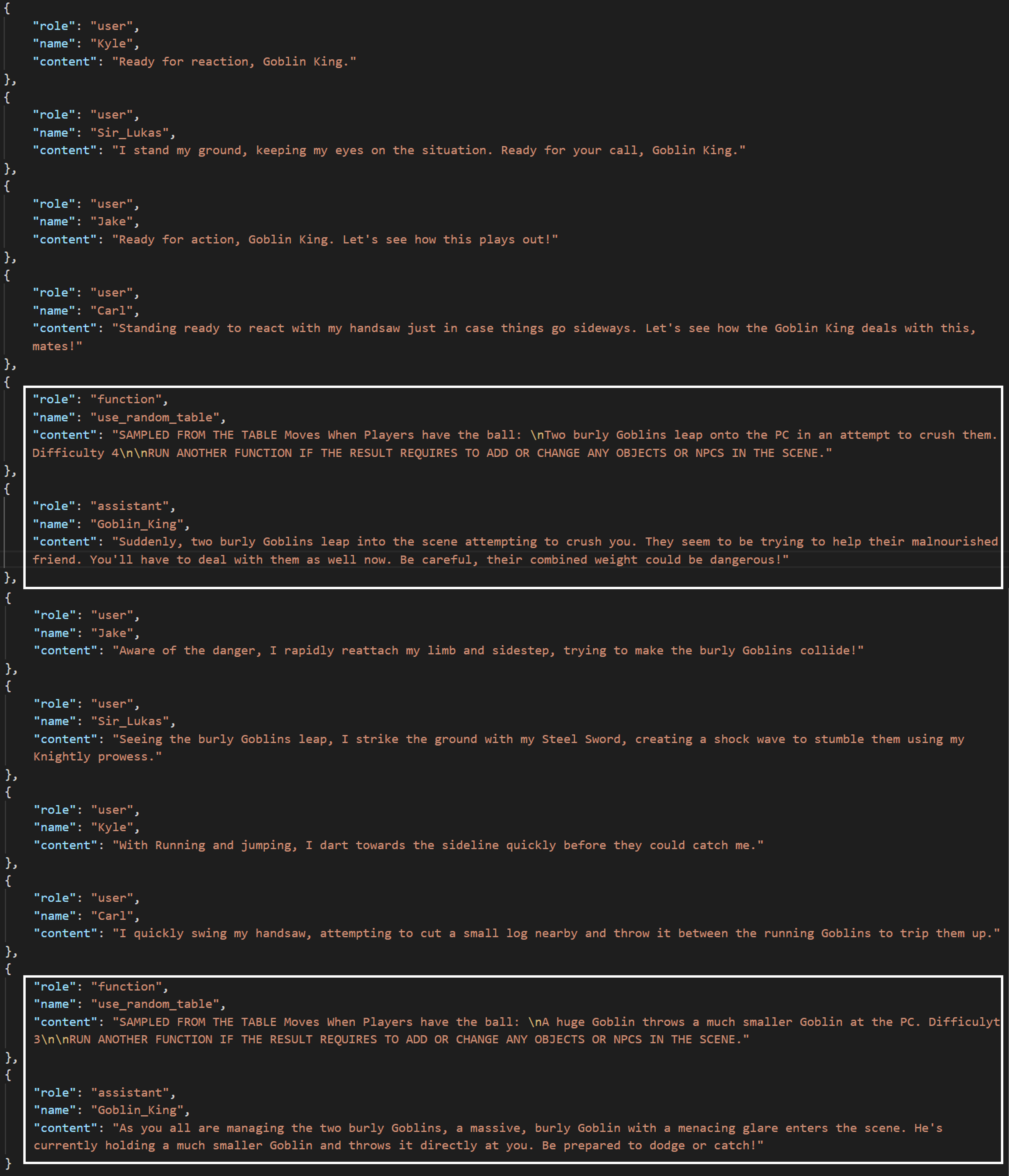} \hfill
  \caption {Without the dice roll function, one of the state functions, {\tt use\_random\_table}, is called without resolving the previous challenge, introducing a new context again.}
\end{figure}

\onecolumn
\subsubsection{Accepting the requests too generously only with a dice roll function}

\begin{figure}[H]
  \includegraphics[width=1.0\linewidth]{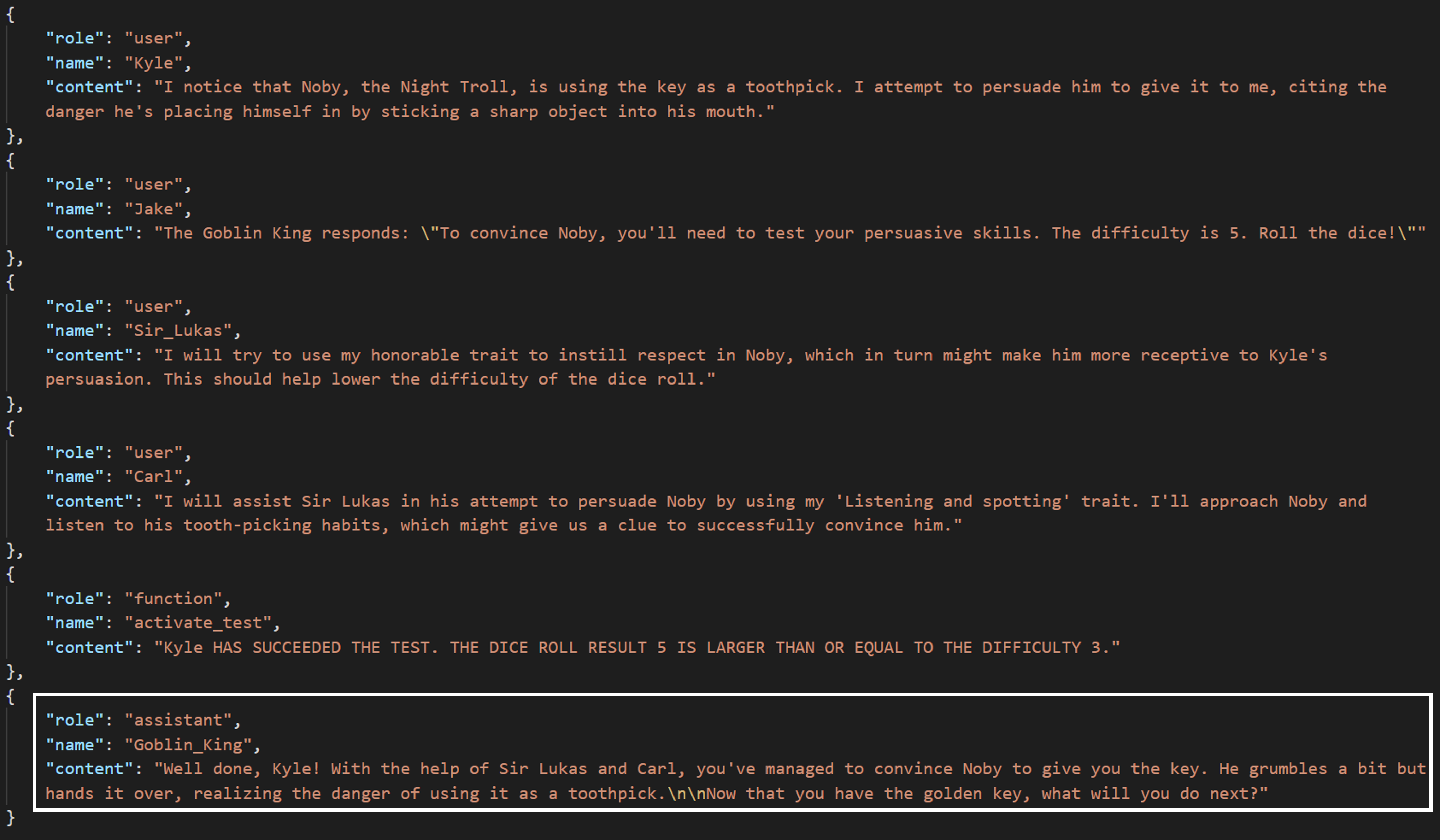} \hfill
  \caption {The evaluator gave the lowest score to this response, since he thought Carl's help was completely ridiculous but the game master called the dice roll function without validating it. Also, it failed to point out Jake's invalid behavior.}
\end{figure}

\onecolumn
\section{Unit test example}
\label{sec:appendixI}

\begin{figure}[H]
  \includegraphics[width=1.0\linewidth]{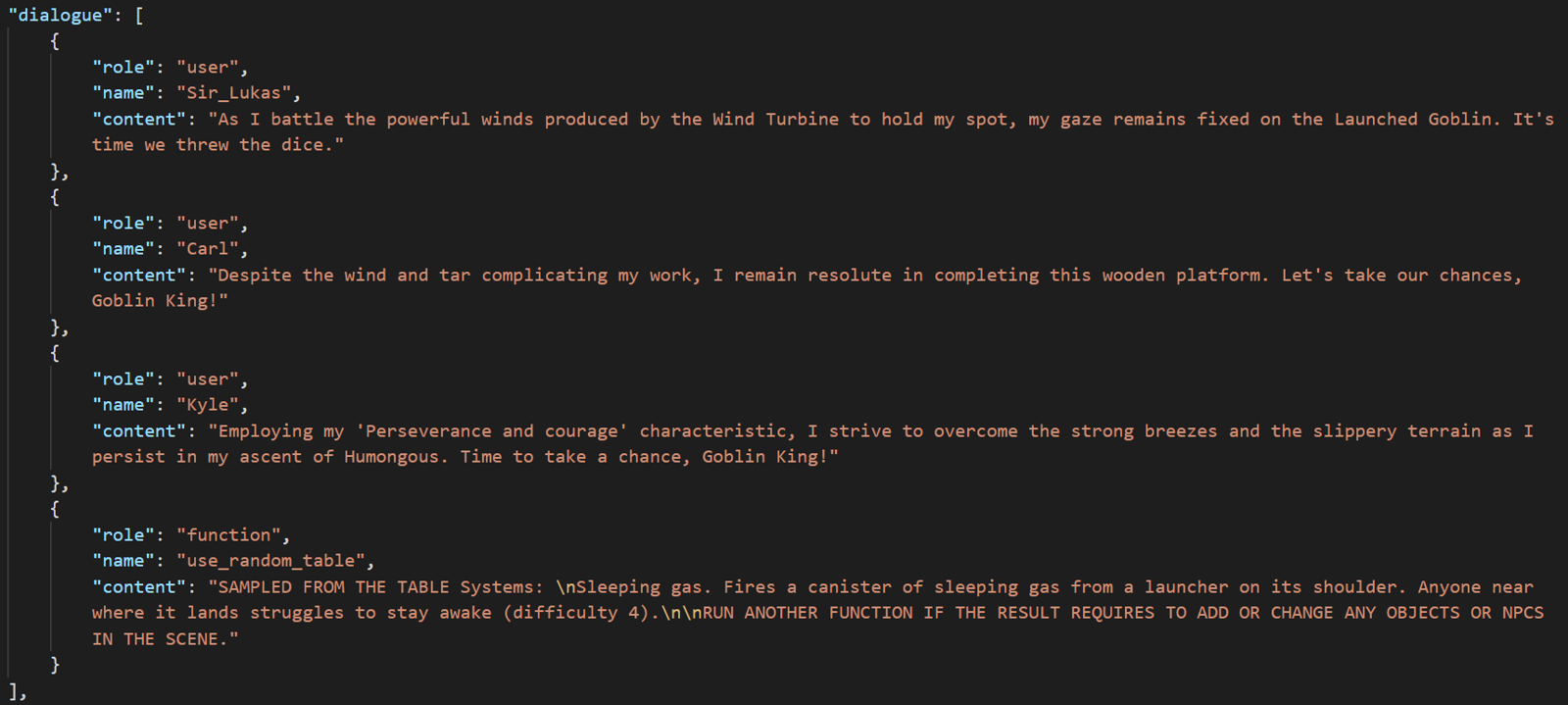} \hfill
  \caption {A sample dialogue in one of the unit tests. Since a new object "Sleeping gas canister" has been sampled from a random table, it should be added to the scene via {\tt add\_object} function.}
\end{figure}

Figure 12 shows a sample dialogue from one of the test cases we crafted. In this dialogue, a humongous fired sleeping gas canister, which has been sampled from a random table. To pass this test case, the game master should call \verb|add_object| function to add a new object "Sleeping gas canister".

Figure 13 shows the result after feeding the input dialogue into \textbf{FG-all}. We can see that \verb|activate_test| function was called multiple times, but the function \verb|add_object| was not, on the other hand. This can be helpful in terms of the actual gameplay, since it allows the players to resolve the given challenge via dice rolls. However, in terms of the game state update, it is considered as wrong since the object has not been added to the scene.

On the other hand, \textbf{FG-states} correctly called \verb|add_object| function as intended, which is shown in Figure 14. Since there is no dice roll function to call, the API could fetch a proper state function without distraction. Since it successfully added the new object, it passed this test case. However, as we saw from the human evaluation results, this is not always beneficial in terms of the game's progress.

\onecolumn

\begin{figure}[H]
  \includegraphics[width=1.0\linewidth]{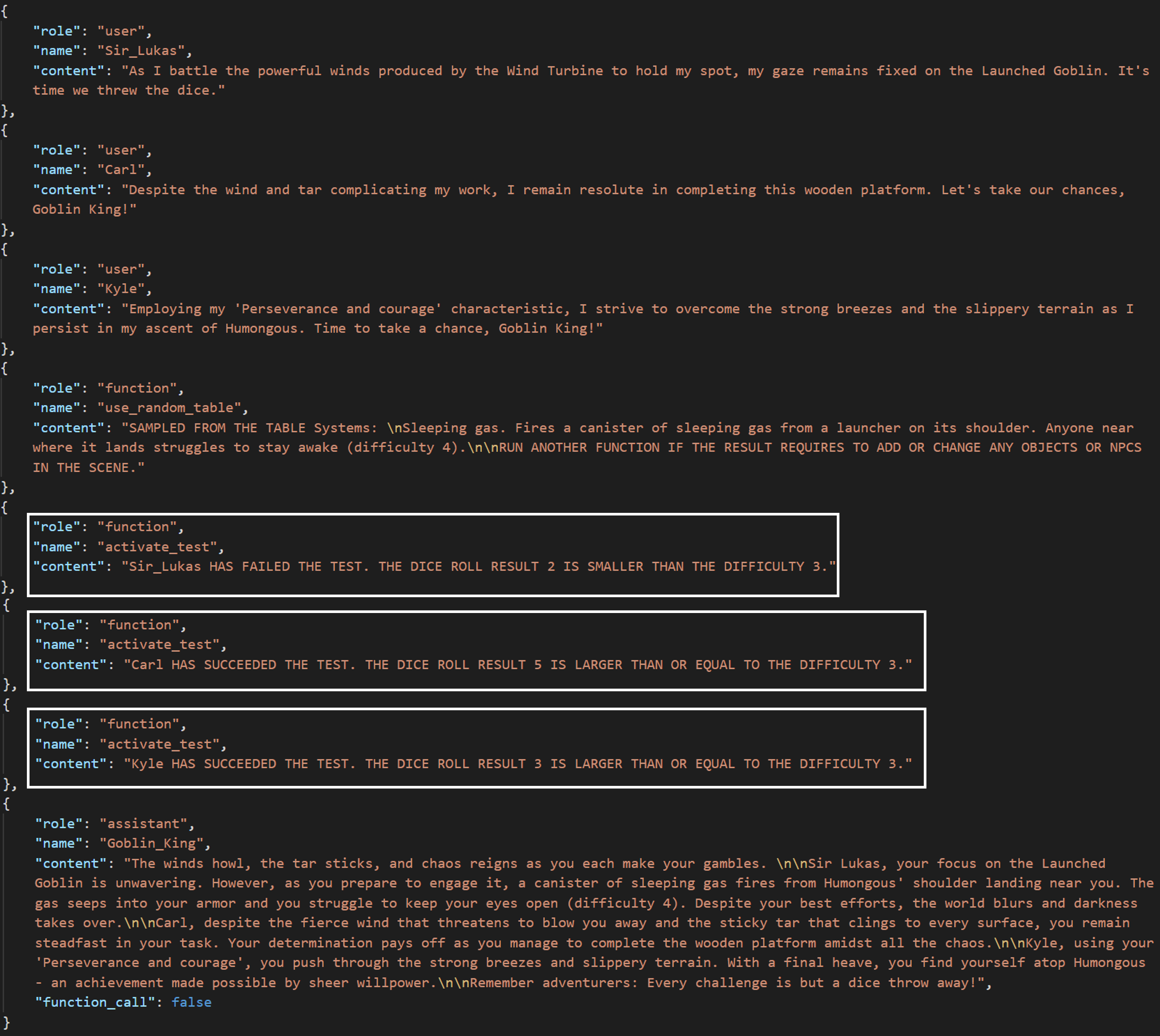} \hfill
  \caption {Generated responses and function call results from \textbf{FG-all}. The dice roll function was called multiple times and the master didn't call {\tt add\_object} function.}
\end{figure}

\begin{figure}[H]
  \includegraphics[width=1.0\linewidth]{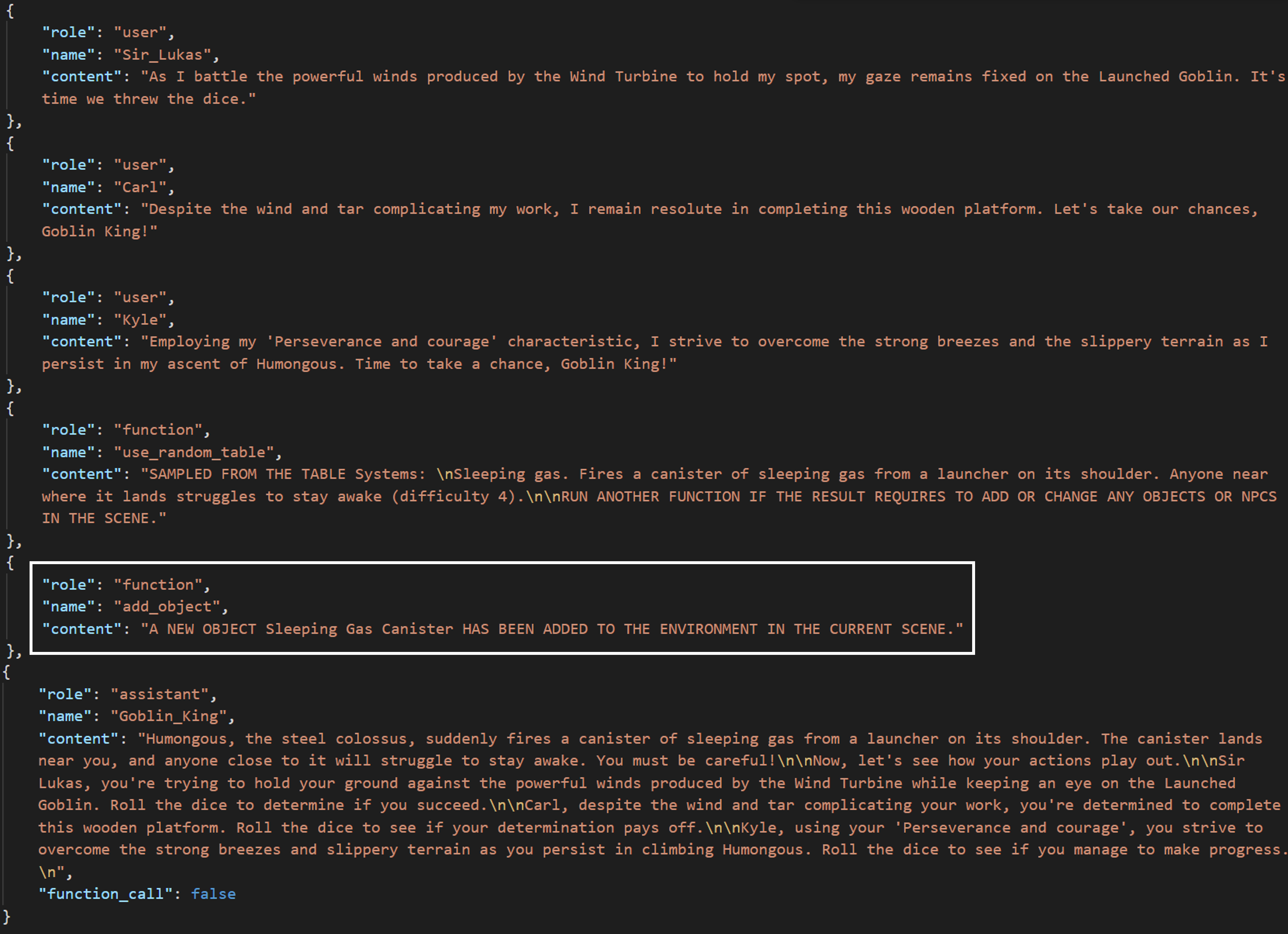} \hfill
  \caption {Generated responses and function call results from \textbf{FG-states}. Without a dice roll, {\tt add\_object} function was correctly called.}
\end{figure}

\end{document}